\documentclass[10pt,twocolumn,letterpaper]{article}

\usepackage[pagenumbers]{wacv} 

\usepackage{xspace}
\newcommand{\algname}{\textsc{Gen-AFFECT}\xspace}
\usepackage{afterpage}

\definecolor{wacvblue}{rgb}{0.21,0.49,0.74}
\usepackage[pagebackref,breaklinks,colorlinks,allcolors=wacvblue]{hyperref}

\def\wacvPaperID{1003} 
\def\confName{WACV}
\def\confYear{2026}

\title{Gen-AFFECT: Generation of Avatar Fine-grained Facial Expressions with Consistent identiTy}

\author{Hao Yu\\
Boston University\\
Boston, MA
\and
Rupayan Mallick\\
Georgetown University\\
Washington, D.C.
\and
Margrit Betke\\
Boston University\\
Boston, MA
\and
Sarah Adel Bargal\\
Georgetown University\\
Washington, D.C.
}

\begin{document}
\twocolumn[{%
\renewcommand\twocolumn[1][]{#1}%
\maketitle
  \centering
  \includegraphics[width=0.8\textwidth]{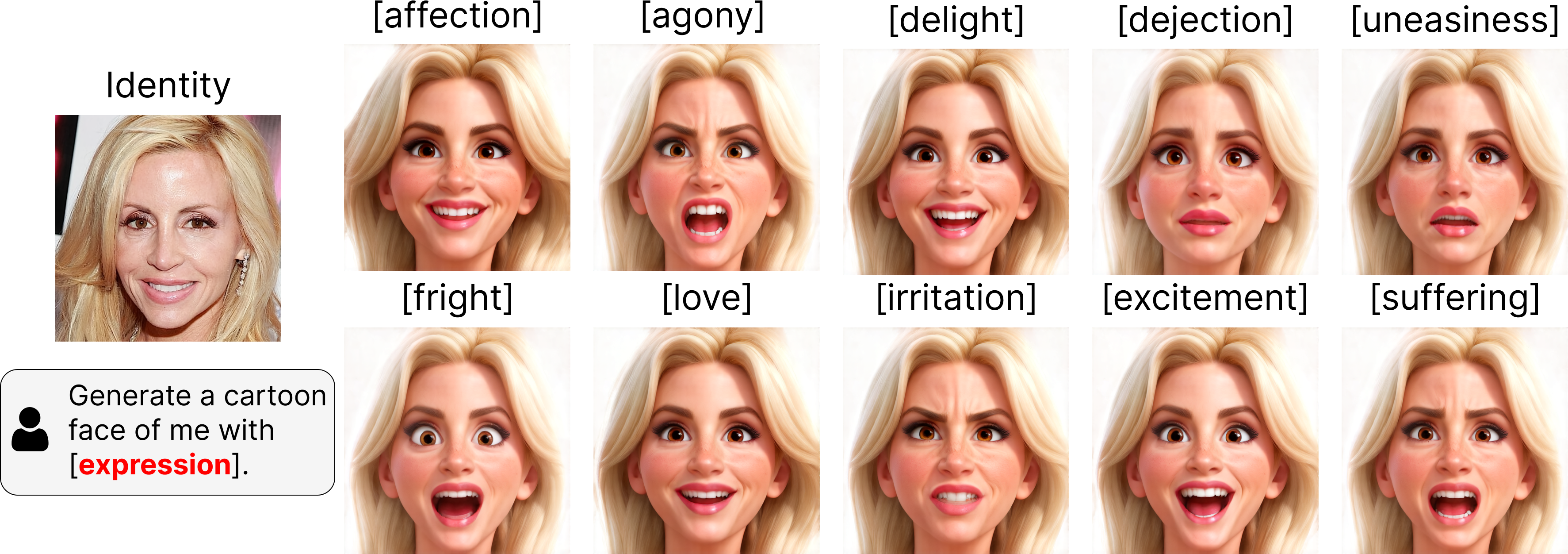}
  \captionof{figure}{The proposed framework, \algname, generates high-quality
  2D personalized avatars representing a given identity with fine-grained facial expressions and consistent identity.}
\vspace{1em}
\label{fig:teaser}
}]

\begin{abstract}
Different forms of customized 2D avatars are widely used in gaming applications, virtual communication, education, and content creation. However, existing approaches often fail to capture fine-grained facial expressions and struggle to preserve identity across different expressions. We propose \textbf{\algname}, a novel framework for personalized avatar generation that generates expressive and identity-consistent avatars with a diverse set of facial expressions. Our framework proposes conditioning a multimodal diffusion transformer on an extracted identity-expression representation. This enables identity preservation and representation of a wide range of facial expressions. \algname additionally employs consistent attention at inference for information sharing across the set of generated expressions, enabling the generation process to maintain identity consistency over the array of generated fine-grained expressions. \algname demonstrates superior performance compared to previous state-of-the-art methods on the basis of the accuracy of the generated expressions, the preservation of the identity and the consistency of the target identity across an array of fine-grained facial expressions. 

\end{abstract}
    
\section{Introduction}
\label{sec:intro}
Personalized 2D avatars have gained increasing interest due to their wide range of applications in gaming~\cite{nizam2022avatar}, virtual communication~\cite{panda2022alltogether}, and education~\cite{fink2024ai,segaran2021does}. For example, avatars can serve as digital representations for users in virtual environments, offering a personalized and expressive way of communication. A key requirement in such scenarios is to generate avatars that not only resemble a specific individual but also convey a variety of facial expressions. However, existing methods for avatar generation often suffer from limited expressivity, typically constrained to a small set of basic emotions such as happiness or sadness. They also struggle to preserve a consistent identity across different expressions, producing variations in facial features. The ability to produce fine-grained, identity-consistent expressions remains a challenge.

Recent advances in text-to-image (T2I) generative models such as Stable Diffusion~\cite{rombach2022high}, DALLE~\cite{betker2023improving}, and Imagen~\cite{saharia2022photorealistic} have enabled the generation of high-quality images from text prompts. Building on this progress, recent work such as DreamBooth~\cite{ruiz2023dreambooth} and Textual Inversion~\cite{gal2023an} has explored the personalization of T2I models, where, given a reference image or a set of reference images, the model generates subject-driven content that preserves visual identity across various text prompts. Among these, face personalization has emerged as a prominent direction due to its broad human-centered applications. Methods like PhotoMaker~\cite{li2024photomaker} and InstantID~\cite{wang2024instantid} preserve face identity by conditioning T2I models on face embeddings, allowing personalized facial image generation guided by text prompts.

Inspired by the success of face personalization in realistic domains, we propose a novel framework, \algname, for avatar generation that enables both identity control and fine-grained expression synthesis. Our method leverages a state-of-the-art T2I model, Stable Diffusion 3.5 (SD 3.5)~\cite{esser2024scaling}, and integrates both an identity image and an expression exemplar as conditioning input. The proposed framework is capable of generating expressive, identity-consistent avatars for a given identity across a wide range of facial expressions.

In this work, we propose an identity-expression conditioned multimodal diffusion transformer specifically designed for controllable avatar generation. Given a target expression word, an exemplar expression image is retrieved from a curated expression dataset~\cite{chen2022semantic}. The identity and expression embeddings are then extracted using a pretrained face encoder and an expression encoder, respectively. These embeddings are then injected into the multimodal diffusion transformer via decoupled cross-attention~\cite{ye2023ip-adapter}, using an additional set of attention weights to enable effective integration of both identity and expression information.
We train the model using a combination of rectified flow loss, identity loss, and expression loss to ensure faithful expression generation and identity preservation.

In addition, we incorporate a consistent attention mechanism for consistent generation across different facial expressions. Consistent attention~\cite{zhou2024storydiffusion} enables information sharing between the samples within a batch for consistent identity and appearance of generated images. When generating a batch of images across different expressions, it samples reference tokens from the features of other images in the batch and incorporates them into the current attention computation. This in-batch interaction encourages alignment in identity, appearance, and attire. Our model thus can generate diverse facial expressions of a given identity while maintaining consistency in appearance and identity.

Extensive experiments have been conducted to evaluate the proposed framework in terms of expression accuracy, identity preservation, and consistency. Both quantitative and qualitative results show that our method outperforms existing approaches across all three aspects. Additionally, we present qualitative results in alternative artistic styles, highlighting the flexibility of the proposed framework. We summarize our contributions as follows:
\begin{itemize}
    \item We propose \algname, a framework to generate identity-consistent avatars with fine-grained facial expressions.
    \item \algname is able to achieve identity-expression control by embedding identity and expression into a multimodal diffusion transformer. 
    \item \algname is able to achieve identity-consistency over an array of generated facial expressions by information sharing across generations at inference time.
    \item Experimental results show superior expression accuracy, identity preservation, and consistency achieved by  \algname compared to existing approaches.
\end{itemize}

\section{Related Work}

This work represents a form of face personalization and is also related to facial expression generation. 

\subsection{Text-to-Image Face Personalization}
In a general context, personalization using Text2Image models predominantly involves using a textual description to edit a given object in a different setting ~\cite{ruiz2023dreambooth, ye2023ip-adapter, gal2023an}. Text2Image face personalization focuses on generating identity-preserving personalized faces from text descriptions, building on the success of diffusion-based text-to-image models. DreamBooth~\cite{ruiz2023dreambooth} and Textual Inversion~\cite{gal2023an} are two foundational approaches that achieve subject-driven generation by fine-tuning a special prompt token to represent the target identity. These two foundational approaches fail to preserve the identity when applied to faces, and also require multiple images for the same identity. Some of the identity-preserving approaches include FaceStudio~\cite{yan2023facestudio}, FastComposer~\cite{Xiao_fastcomposer_ijcv}, PhotoMaker~\cite{li2024photomaker}, PortraitBooth~\cite{peng2024portraitbooth}, InstantID~\cite{wang2024instantid}, MetaPortrait~\cite{Zhang_2023_CVPR}. Methods like PhotoMaker~\cite{li2024photomaker} use multiple images to extract identity features and then inject the stacked features, which might not always be feasible and also limit the method for real-time applications. Other methods such as PortraitBooth~\cite{peng2024portraitbooth} inject facial/identity feature embeddings using a facial feature extractor. InstantID~\cite{wang2024instantid} uses IP-Adapter~\cite{ye2023ip-adapter} and ControlNet~\cite{zhang2023adding} as the backbone of the model. MetaPortrait~\cite{Zhang_2023_CVPR} uses a meta learning based approach for indentity based portrait generation. The IP-Adapter~\cite{ye2023ip-adapter} incorporated face ID embeddings through decoupled cross-attention modules for identity-preserving generation.
Other methods like Face0~\cite{valevski2023face0}, PortraitBooth~\cite{peng2024portraitbooth}, Face2Diffusion~\cite{shiohara2024face2diffusion}, and PhotoMaker~\cite{li2024photomaker} inject identity information by replacing or fusing face embeddings extracted via CLIP~\cite{radford2021learning} or a face recognition model with text embeddings, allowing diffusion models to condition on identity embeddings. 

\subsection{Facial Expression Generation}

Facial expression generation aims to synthesize or manipulate realistic facial expressions in images or videos. 
The early methods predominantly relied on Generative Adversarial Networks (GANs)~\cite{goodfellow2020generative}. Few of the works aimed at editing or manipulating the expressions include StarGAN~\cite{choi2018stargan}, GANimation~\cite{ganimation}, GANMut~\cite{d2021ganmut}, ICFace~\cite{Tripathy_2021_WACV}, Neural Emotion Director~\cite{paraperas2022ned}. 
Notably, StarGAN~\cite{choi2018stargan} introduced a unified framework for multi-domain facial attribute transfer, enabling basic expression editing. GANMut~\cite{d2021ganmut} employs a continuous expression control spaces such interpretable conditional space to condition the model. 
Subsequent works aimed to obtain finer control over expression generation by leveraging Facial Action Units (AUs)~\cite{pumarola2018ganimation,tripathy2020icface, zhao2021action}, and a valence-arousal space~\cite{azari2024emostyle}.

Recently, diffusion models~\cite{ho2020denoising} have emerged as state-of-the-art in image generation due to their superior sample quality and diversity. 
Stable Diffusion~\cite{rombach2022high}, a powerful latent diffusion model, demonstrated remarkable abilities in text-to-image generation, including limited expression synthesis via text prompts. Building on this, Pikoulis \textit{et al.}~\cite{pikoulis2023photorealistic} adapted Stable Diffusion to generate seven basic expressions using CLIP-based latent guidance. In addition,~\cite{kaifeng_4d_Diffusion} uses diffusion models to generate facial expressions and condition on 3D meshes. 
More relevant to our work, Liu \textit{et al.}~\cite{liu2024towards} introduced a face generation framework capable of identity-expression control and fine-grained expression synthesis through a conditional diffusion model. In our work, in addition to addressing identity-expression control, we additionally enable identity-consistent generations of arrays of fine-grained facial expressions.





\begin{figure*}[t]
    \centering
    \includegraphics[width=0.85\textwidth]{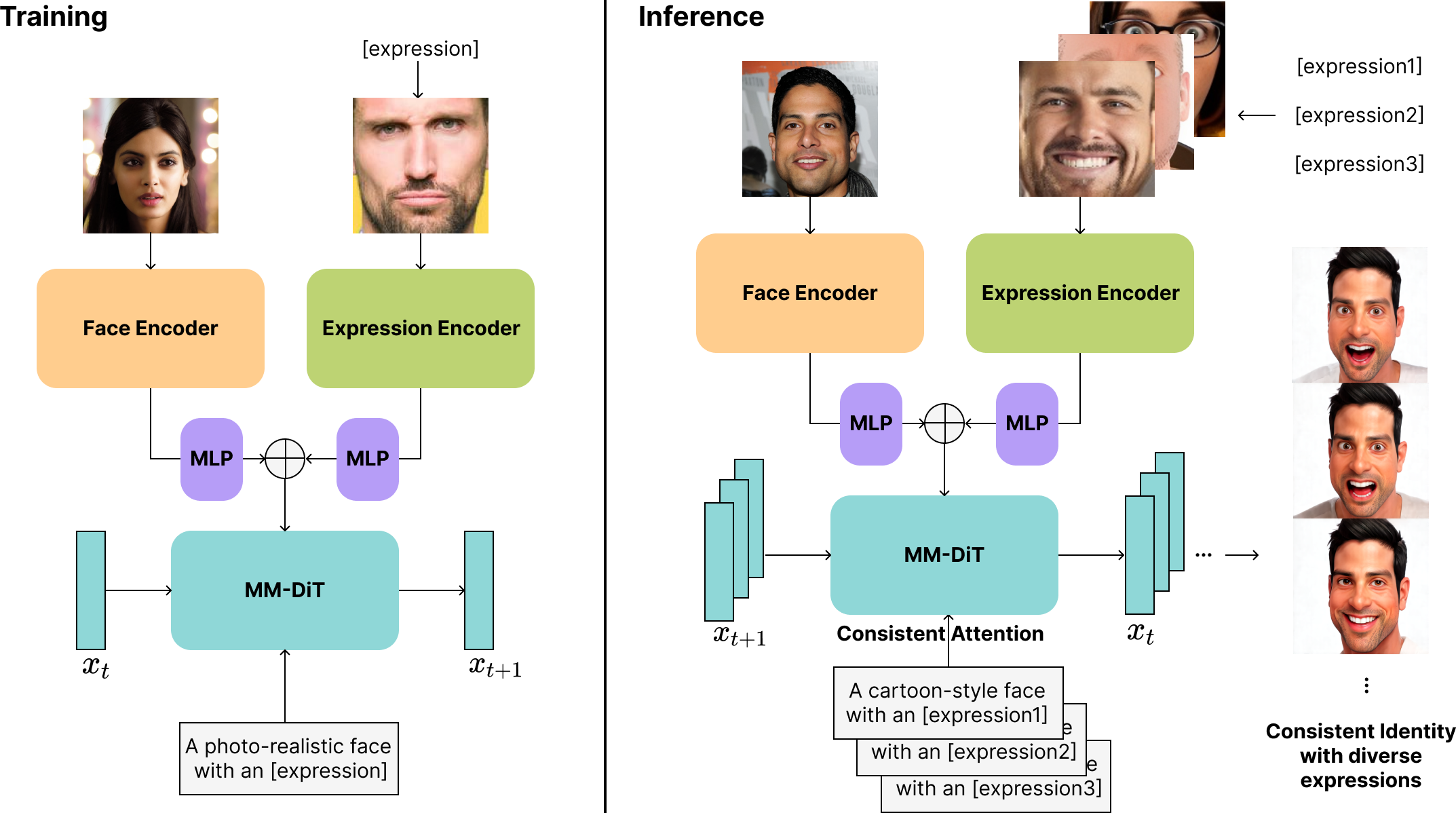}
    \caption{Overview of the proposed framework. \algname includes a face encoder for extracting the identity embedding, an expression encoder for capturing the expression embedding, MLPs to project both embeddings, and MM-DiT—a multimodal diffusion transformer performing one diffusion step conditioned on text and the fused identity-expression embedding. The framework has two phases: training, where given an identity image and an expression image retrieved by expression class, the model learns to generate photo-realistic faces; and inference, where a single identity image and a set of expression images produce avatars with consistent identity and diverse expressions via consistent attention in MM-DiT.
    }
    \label{fig:pipeline}
    \vspace{-1em}
\end{figure*}

\section{Method}

\algname consists of a face encoder, an expression encoder, MLPs, and the multimodal diffusion backbone, MM-DiT~\cite{esser2024scaling} (Multimodal Diffusion Transformer) (Figure~\ref{fig:pipeline}). Given a reference identity face image, a target facial expression, and a text prompt, the model learns to generate an image that keeps the reference identity, conveys the desired expression, and aligns with the semantic guidance of the text prompt. \algname retrieves a reference expression image from a large expression database based on the target facial expression class, and conditions the diffusion model on the extracted identity feature, the expression feature and the text prompt. To further improve the consistency of the generated faces, \algname utilizes \textit{Consistent Attention} to generate subject-consistent images with diverse expressions during inference. In general, \algname is capable of generating personalized avatars of the given subject with diverse facial expressions and consistent identity and appearance.

\subsection{Preliminaries}\label{sec:preliminaries}

\noindent\textbf{Rectified Flows.}
We build our framework on Stable Diffusion 3.5 (SD 3.5)~\cite{esser2024scaling}, a state-of-the-art latent text-to-image generative model trained using the Rectified Flow framework. In this formulation, the generative model learns to map samples $x_0$ from a noise distribution $p_0$ to samples $x_1$ from a data distribution $p_1$ in terms of an ordinary differential equation (ODE) defined over time step $t\in[0,1]$:
\begin{equation}
    dx_t = v_\theta(x_t, t)dt,
    \label{eq:velocity_ode}
\end{equation} 
where the velocity field $v_\theta(x_t, t)$ is parameterized by a neural network. 
Directly solving this equation is expensive.
Rectified flows~\cite{liu2023flow,lipman2023flow} define a forward process where distributions $p_0$ and $p_1$ are connected by straight-line trajectories, as given in Equation~\eqref{eq:rectified_flow}:
\begin{equation}
    x_t = (1 - t)x_0 + tx_1, \quad x_0 \sim \mathcal{N}(0, I), \quad t \in [0,1],
    \label{eq:rectified_flow}
\end{equation}  
where $p_0 = \mathcal{N}(0, I)$.  
The network can then be trained by minimizing the mean-squared error between the neural velocity field and the direction of linear trajectories connecting random points sampled from $p_1$ and $p_0$. This direction can be derived based on Equation~\eqref{eq:rectified_flow} as $\frac{dx_t}{dt} = x_1 - x_0$.
The objective is then formulated as:  
\begin{equation}
    \mathcal L_{\text{rf}} =  \mathbb{E}_{t \sim P(t), x_0 \sim p_0, x_1 \sim p_1} 
    \left[ \| v_\theta(x_t, t) - (x_1 - x_0) \|^2 \right],
    \label{eq:rf_loss}
\end{equation}  
where $P(t)$ is a distribution over time $t$.

\noindent\textbf{Multimodal Diffusion Transformer (MM-DiT).}
The multimodal diffusion transformer (MM-DiT)~\cite{esser2024scaling} introduces learnable streams for both image and text tokens, enabling bidirectional information flow between modalities. Unlike approaches that inject fixed text representations via cross-attention, MM-DiT employs separate trainable weights for each domain, leading to superior performance in text-to-image generation. Our framework adopts MM-DiT as the text-to-image backbone and integrates both identity and expression conditions to further guide the generation process.

\subsection{Identity and Expression Conditioning}

To enable the pretrained text-to-image model to generate images that preserve the target identity and expression, we first extract embeddings from a reference identity image and a reference expression image using a face encoder and an expression encoder, respectively. 
For the face encoder, a state-of-the-art face recognition model, ArcFace~\cite{deng2019arcface} is used. A state-of-the-art expression recognition model, POSTER~\cite{zheng2023poster} is employed as the expression encoder.
These specialized encoders are preferred over a generic CLIP image encoder due to their improved capacity to disentangle and capture identity and expression-specific features.
To facilitate domain adaptation and align feature dimensions, identity and expression embeddings are further projected through a lightweight, trainable projection network that consists of two-layer MLPs and Layer Normalization~\cite{ba2016layer}.
Each embedding is transformed into a sequence of $N$ feature tokens (with $N = 4$ in this study), where each token has the same dimensionality as the text embedding.
Finally, the identity and expression embeddings are combined through additive fusion followed by two MLP layers with Layer Normalization, and the resulting embeddings are fed into the MM-DiT model.

To effectively integrate identity and expression embeddings into the pretrained MM-DiT model, we introduce additional attention weights inspired by ~\cite{ye2023ip}.
Specifically, the attention layer in the MM-DiT model~\cite{esser2024scaling} is computed as:
\begin{equation}
\begin{aligned}
Q &= [X_{i} W_{Q_i};\ X_{p} W_{Q_p}], \\
K &= [X_{i} W_{K_i};\ X_{p} W_{K_p}], \\
V &= [X_{i} W_{V_i};\ X_{p} W_{V_p}], \\
\operatorname{Attention}&(Q,K,V) = \operatorname{Softmax}\left( \frac{QK^\top}{\sqrt{d}} \right) V,
\end{aligned}\label{eq:attention_sd3}
\end{equation}
where $X_{p}, X_{i}$ are text prompt and image embeddings from the previous layer, $d$ is the dimension of $Q, K$, and $W_{Q_i}, W_{Q_p}, W_{K_i}, W_{V_i}, W_{V_p}$ are weight matrices. Here, $``\textbf{;}"$ denotes the concatenation operation. We incorporate the identity-expression embeddings $X_{ie}$ by adding a set of weight matrices $W_{K_{ie}}, W_{V_{ie}}$:
\begin{equation}
\begin{aligned}
Q' &= X_{i} W_{Q_i}, \\
K' &= [X_{i} W_{K_i};\ X_{ie} W_{K_{ie}}], \\
V' &= [X_{i} W_{V_i};\ X_{ie} W_{V_{ie}}], \\
\operatorname{Attention}&(Q',K',V') = \operatorname{Softmax}\left( \frac{Q'K'^\top}{\sqrt{d}} \right) V'.
\end{aligned}\label{eq:attention_condition}
\end{equation}
The final output would be
\begin{equation}
   Z = \operatorname{Attention}(Q,K,V) + \alpha \operatorname{Attention}(Q',K',V'),
\end{equation}
where $\alpha$ is a scaling factor that controls the effect of the identity and expression conditioning.

\subsection{Training}

To train the model, we employ a combination of rectified flow loss, identity loss, and expression loss.
The rectified flow loss, described in Equation~\ref{eq:rf_loss}, serves as the primary training objective for image generation.
For improved identity preservation, we define an identity loss as:
\begin{equation}
    \mathcal L_{\text{id}} = 1 - \operatorname{CosSim}\left(\mathcal{E}_{\text{id}}(I_{\text{id}}), \mathcal{E}_{\text{id}}(x_1)\right),
\end{equation}
where $I_{\text{id}}, x_1$ denote the reference image and the generated image, respectively, and $\mathcal{E}_{\text{id}}$ is the pretrained face encoder~\cite{deng2019arcface}. It computes the cosine similarity between the identity embeddings extracted from the reference identity image $I_{\text{id}}$ and the generated image $x_1$.
Maximizing this similarity ensures that the generated image closely preserves the reference identity.
Similarly, for expression consistency, we employ a mean squared error (MSE) loss:
\begin{equation}
    \mathcal L_{\text{exp}} = \operatorname{MSE}\left(\mathcal{E}_{\text{exp}}(I_{\text{exp}}), \mathcal{E}_{\text{exp}}(x_1)\right),
\end{equation}
where $\mathcal{E}_{\text{exp}}$ represents the pretrained expression encoder~\cite{zheng2023poster}. Minimizing this loss encourages the predicted image to faithfully reflect the target expression.
For efficiency, we approximate the generated image using a one-step sampling strategy based on the rectified flow formulation in Section~\ref{sec:preliminaries}:  
\begin{equation}
    x_1 \approx x_t + v_\theta',
\end{equation}
where $x_t$ is a noisy sample at time $t$, and $v_\theta'$ is the learned velocity field through our multimodal diffusion transformer parametrized by $\theta$ (approximated at time step $t$).

The overall training objective is:
\begin{equation}
   \mathcal L = \mathcal L_{\text{rf}} + \beta_1\mathcal L_{\text{id}} + \beta_2\mathcal L_{\text{exp}},
\end{equation}
where $\beta_1, \beta_2$ are scaling factors.

\begin{table*}
    \caption{Quantitative results showing that \algname achieves superior performance compared to prior state-of-the-art across all three dimensions: expression accuracy, identity preservation, and consistency. }
    \label{table:result_main}
    \centering
    \normalsize
    \begin{tabular}{lcccccc}
        \hline
         & \multicolumn{2}{c}{Expression} & \multicolumn{2}{c}{Identity} & \multicolumn{2}{c}{Consistency} \\
         Model &  Exp.$\downarrow$ & CLIP$\uparrow$ & ID.$\uparrow$ & DINO$\uparrow$ & DINO Con.$\uparrow$ & ID Con.$\uparrow$ \\
        \hline
    FastComposer~\cite{Xiao_fastcomposer_ijcv} & 14.17 & 0.615 & 0.253 & 0.820 & 0.883 & 0.506 \\
    PuLID~\cite{guo2024pulid} & 13.58 & 0.642 & 0.210 & 0.766 & 0.945 & 0.628 \\
    PhotoMaker~\cite{li2024photomaker} & 13.39 & 0.661 & 0.103 & 0.621 & 0.934 & 0.572 \\
    Conditional SDXL & 11.57 & 0.667 & 0.220 & 0.803 & 0.921 & 0.488 \\
    \algname (Ours) & \textbf{11.09} & \textbf{0.678} & \textbf{0.361} & \textbf{0.828} & \textbf{0.957} & \textbf{0.762} \\
        \hline
    \end{tabular}
\end{table*}

\subsection{Inference with Consistent Attention}

\algname is trained to generate images that preserve both the identity and facial expression of the input reference images. 
During inference, the model takes as input a single reference identity image and a set of target expression classes selected from a predefined expression dictionary comprising 135 categories~\cite{shaver1987emotion}, which semantically capture a broad spectrum of emotional states. For each expression class, the model retrieves a reference image from the large-scale expression database, Emo135~\cite{chen2022semantic}. These reference images serve as expression conditioning signals to guide the generation process.

To ensure consistency of identity and appearance across all generated images with different expressions, we employ a consistent attention mechanism~\cite{zhou2024storydiffusion} that enables shared information within a batch. Specifically, when computing the attention as in Equation~\ref{eq:attention_sd3}, we allow token-level interaction between each image in a batch. Formally, let $X_i^j$ denote the image embedding of the $j$-th image in a batch of size $B$. To encourage cross-image alignment, we randomly sample $S_i^j$ from the features of the remaining images:
\begin{equation}
S_i^j = \operatorname{Samp}(\{X_i^k \mid k \in {1, \dots, B\} \setminus \{X_i^j\}}).
\end{equation}
The sampled image embedding $S_i^j$ is concatenated with the original image embedding $X_i^j$ to form an augmented embedding $F_i^j$.
We then compute key and value projections on $F_i^j$, denoted as $F_i^jW_{K_i}$ and $F_i^jW_{V_i}$, respectively, while retaining the original query $X_i^jW_{Q_i}$. The attention operation is then performed as described in Equation~\eqref{eq:attention_sd3}, where the weights and other features remain the same and only the image embeddings are updated.
This mechanism allows each image to integrate global batch-level context, which encourages convergence in attributes such as character identity, facial features, and artistic style across expressions, improving coherence in generated images.

Further, the model is capable of generating images in a cartoon-like style by leveraging the expressive power of text prompts. Specifically, during inference, users can specify stylistic instructions such as "in cartoon style", "anime rendering", or "Pixar-like character" as part of the text input. These prompts are encoded by the pretrained text encoder and integrated into the diffusion process, guiding the model to synthesize images that adhere not only to the target identity and expression, but also to the desired artistic style. This flexible control enables the model to generalize beyond realistic face synthesis and adapt to stylistic domains such as cartoons or animations.

\section{Experiments}

\subsection{Experimental Setup}

\noindent\textbf{Training Data.}
\algname is trained on the CelebA-HQ dataset~\cite{karras2017progressive}, a high-quality face dataset consisting of 30,000 face images at 1024x1024 resolution. It is widely used for training face generative models. We randomly split the dataset and reserved 100 identities for evaluation. 

\noindent\textbf{Implementation Details.}
Our framework is based on the pretrained Stable Diffusion 3.5 model from Hugging Face.\footnote{https://huggingface.co/stabilityai/stable-diffusion-3.5-large} 
We trained the model for 515k steps with a batch size of 4 in 512×512 resolution. 
Only the projection layers and the added attention weights are trained while other modules are frozen.
The learning rate was set to 1e-4. The weights of identity loss ($\beta_1$) and expression loss ($\beta_2$) were both 0.1.
The AdamW optimizer~\cite{loshchilov2018decoupled} was used with a weight decay of 1e-2.
To enable classifier-free guidance, we randomly drop the text and image conditioning with a probability of 0.05 each, and drop both simultaneously with a probability of 0.05.
The number of inference steps was 50, and the guidance scale was 5.0. The conditioning factor $\alpha$ was set to 0.5.
All experiments were conducted on two NVIDIA RTX A6000 GPUs.

\begin{figure}[t]
    \centering
    \includegraphics[width=\columnwidth]{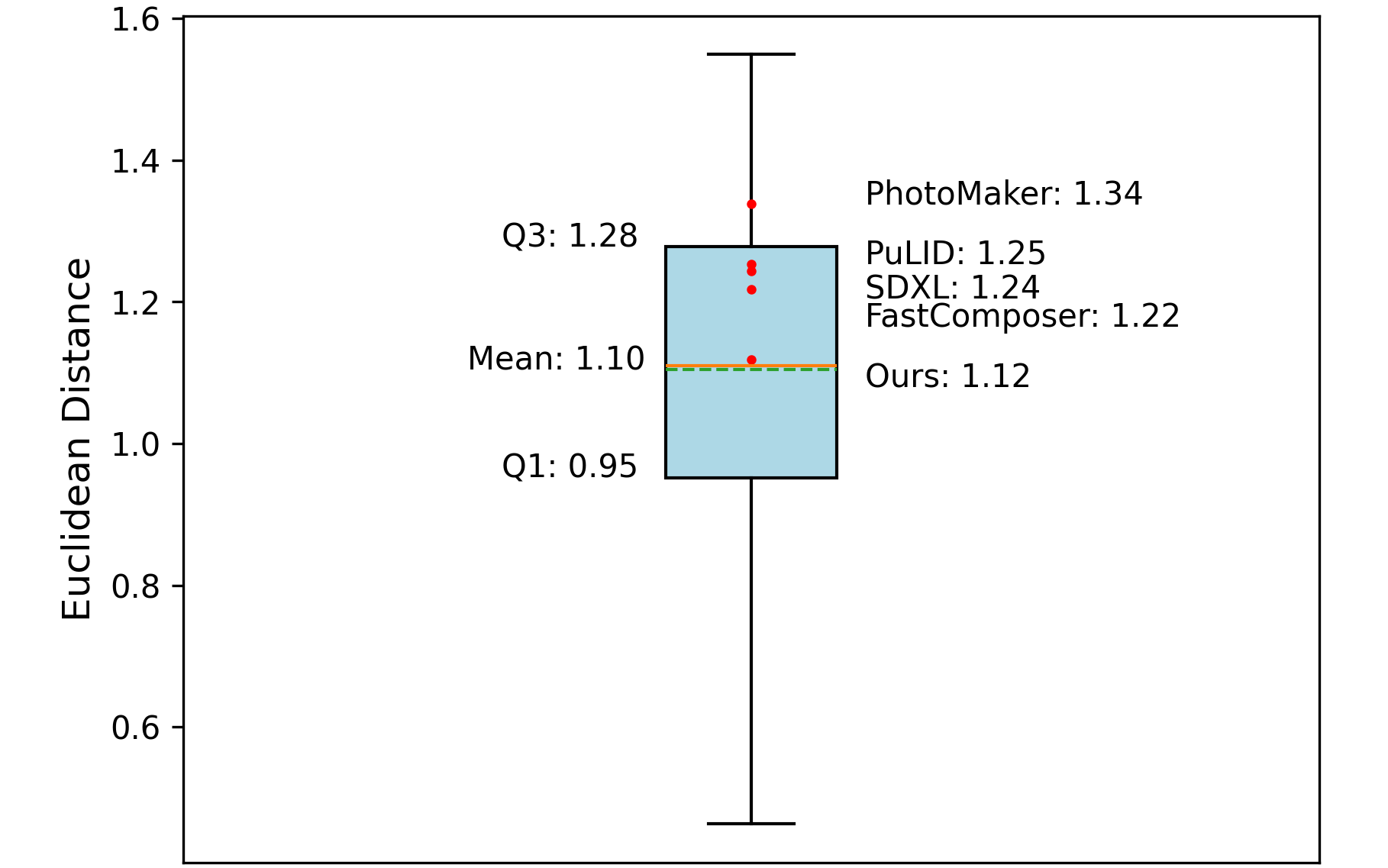}
    \caption{Box plot showing the distribution of pairwise identity distances between real images of the same person for 100 test identities. Each red dot indicates the average identity distance between the model-generated avatars and the reference identity image for each subject. This shows that our generated faces are as identity-consistent as real images of the same person and also better than all other baselines.}
    \label{fig:id_box_plot}
    \vspace{-1em}
\end{figure}

\begin{figure*}[t]
    \centering
    \includegraphics[width=0.8\textwidth]{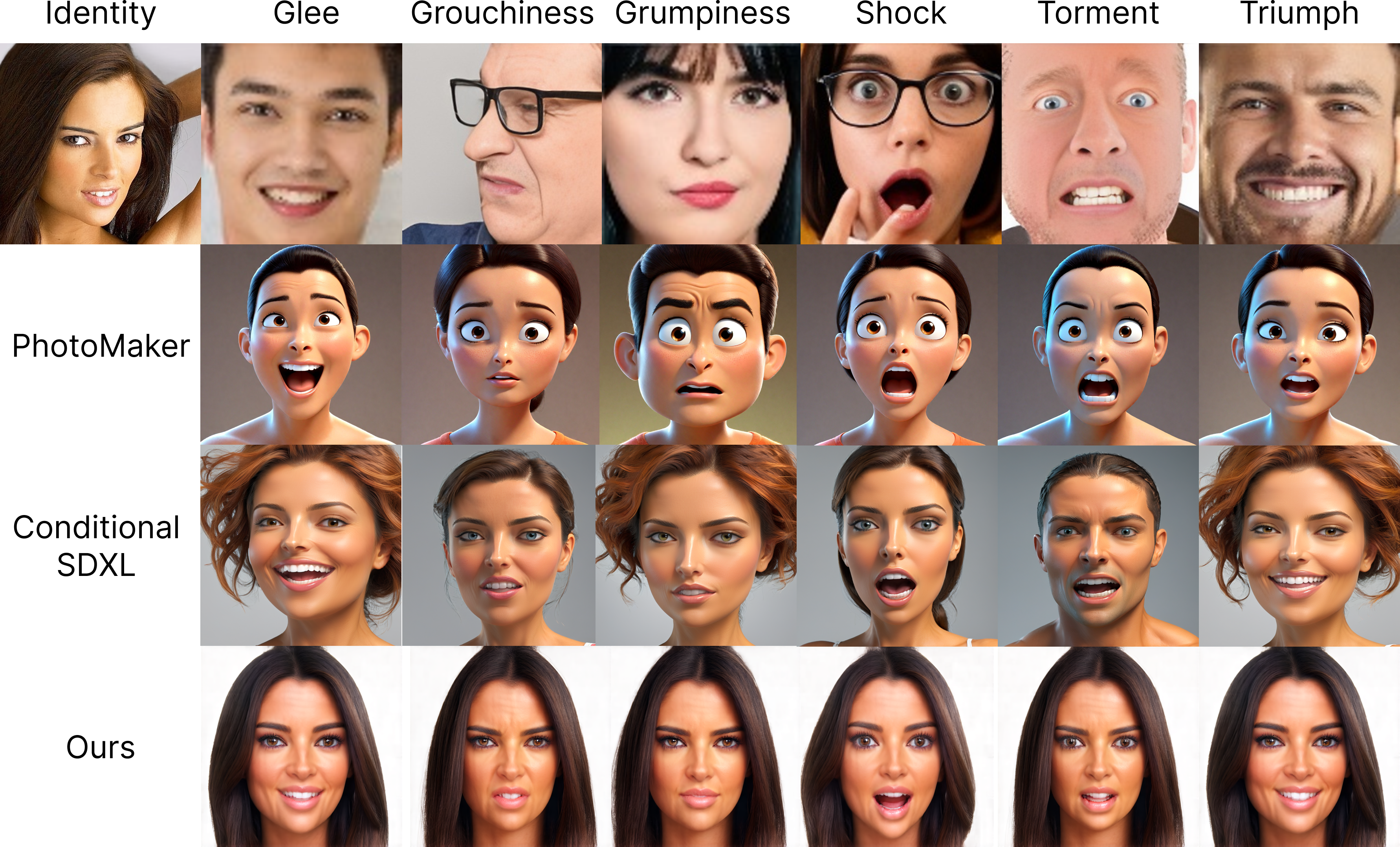}\\
    \vspace{1em}
    \hrule
    \vspace{0.6em}
    \includegraphics[width=0.8\textwidth]{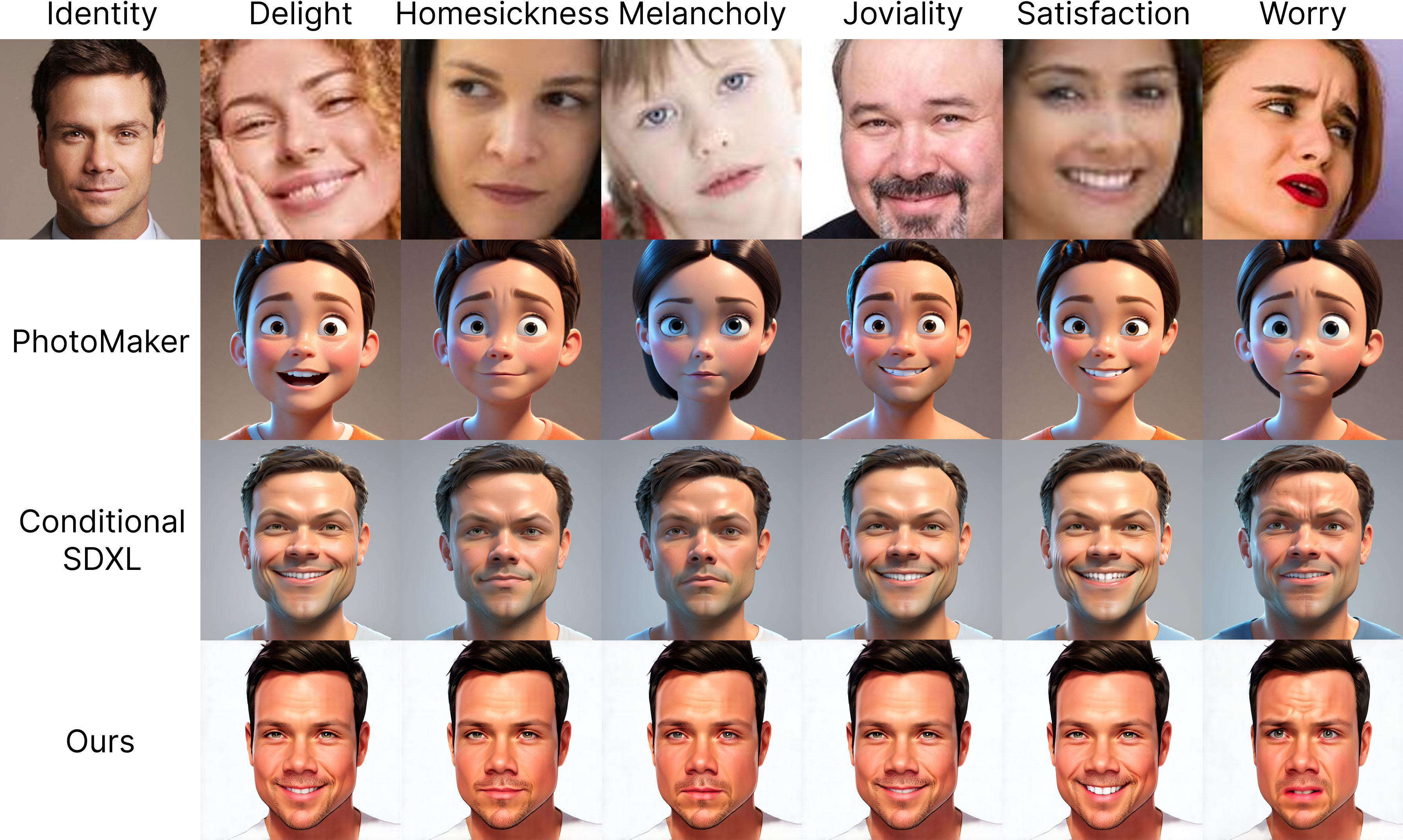}
    
    \caption{Comparison of generated faces from the baseline and the proposed framework. Using the same identity image, six different expressions are generated and compared. The top row shows the reference expression image, retrieved randomly based on the corresponding expression class. Compared to the baseline, our method produces (1) more consistent identity across expressions, (2) more accurate expressions aligned with the reference images, and (3) avatars more similar to the input identity image.
    }
    \label{fig:qualitative_comp}
    \vspace{-1em}
\end{figure*}

\begin{figure*}[t]
    \centering
    \rotatebox{90}{\hspace{0.4cm}\textbf{Avatar}\hspace{0.6cm} \textbf{Identity}}\includegraphics[width=0.8\textwidth]{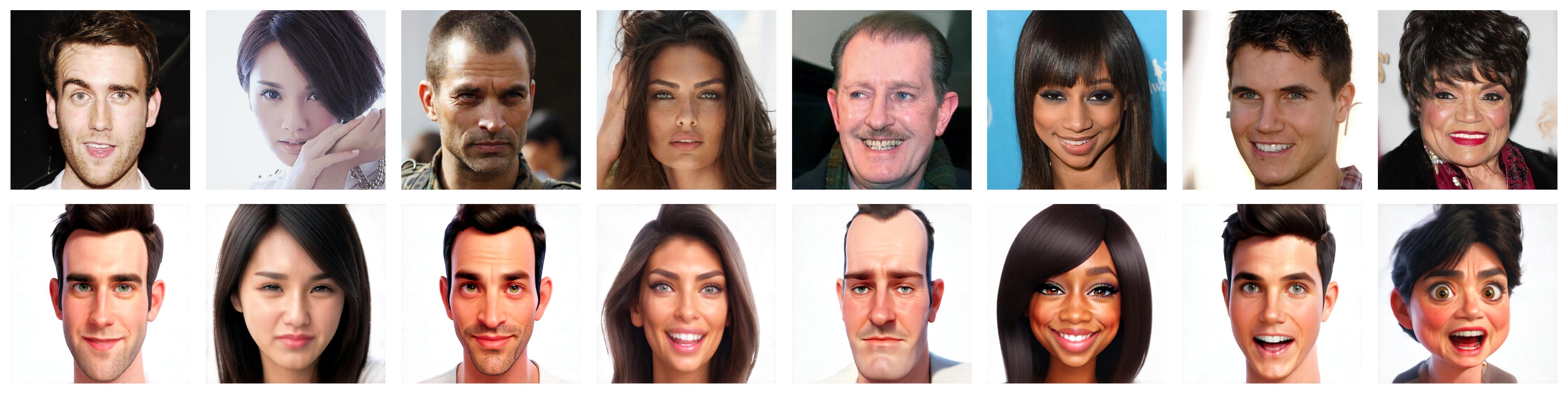}
    \caption{Examples of reference identities and their corresponding generated avatars. Expressions are randomly selected. \algname demonstrates preservation of the key visual features of each input identity.
    }
    \label{fig:id}
\end{figure*}

\begin{figure*}[t]
    \centering
    \textbf{\hspace{2.6cm}Homesickness\hspace{0.3cm} Insecurity\hspace{0.6cm} Joviality\hspace{0.6cm} Melancholy\hspace{0.6cm} Revulsion\hspace{0.6cm} Uneasiness}
    \rotatebox{90}{\hspace{0.6cm}\textbf{Crayon}\hspace{0.6cm} \textbf{Claymation} \hspace{0.6cm} \textbf{LEGO}}\includegraphics[width=0.9\textwidth]{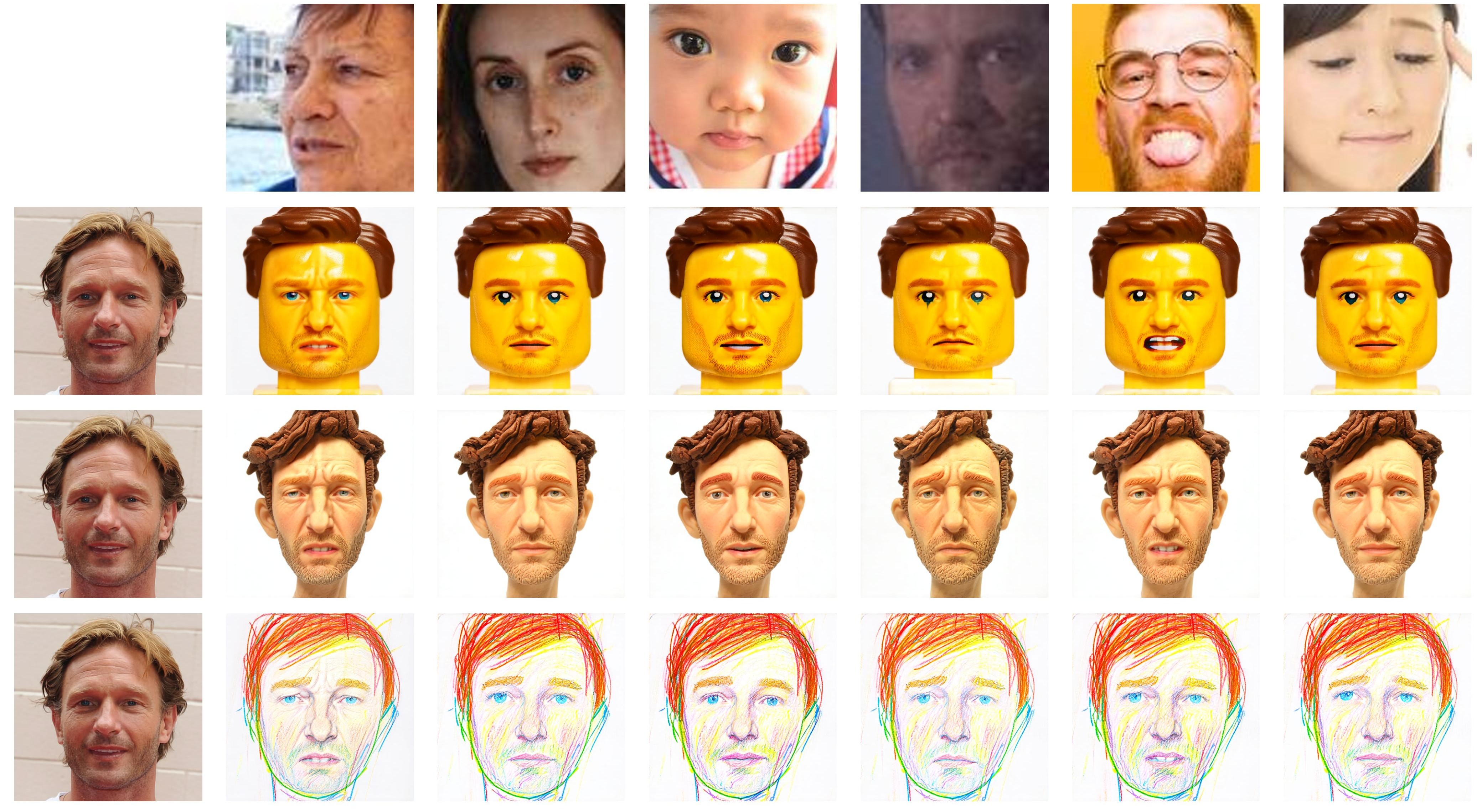}
    \caption{Examples of alternative generation styles: LEGO, claymation, and crayon. These styles are achieved by incorporating curated, style-specific prompts, \textit{e.g.} ``A LEGO-style image of a face showing a facial expression of [expression]. A playful, modular style using brightly colored, interlocking plastic bricks and simple, blocky characters with minimal detail."
    }
    \vspace{-1em}
    \label{fig:style}
\end{figure*}

\noindent\textbf{Evaluation Metrics.}
We evaluate \algname across three dimensions: expression accuracy, identity preservation, and identity consistency across generated expressions.

For expression evaluation, we report expression error (Exp.) and CLIP score (CLIP). Expression error is computed as the average pairwise Euclidean distance between expression embeddings extracted from the generated and reference expression images, using the POSTER model~\cite{zheng2023poster}. Lower values indicate better expression matching. CLIP~\cite{radford2021learning} score measures the average pairwise cosine similarity between CLIP embeddings of the generated and reference expression images, capturing high-level semantic alignment between facial expressions.

For evaluating identity alignment, we use the DINO score and identity similarity (ID.). DINO~\cite{caron2021emerging,oquab2023dinov2} computes the average cosine similarity between DINOv2 embeddings of the generated and reference identity images.
DINO is trained with a self-supervised objective that preserves instance-level details, making it well-suited and widely used for evaluating subjects that could have highly similar text descriptions (e.g., faces with different identities)~\cite{ruiz2023dreambooth}. 
Identity similarity (ID.) is computed using a pretrained face recognition model~\cite{deng2019arcface} and measures the average cosine similarity between the face embeddings of the generated and reference identity images.

To assess identity consistency across generated images, we compute identity consistency (ID Con.) and DINO consistency (DINO Con.), defined as the average pairwise identity similarity and DINO score, respectively, across all 135 generated expression images for a given identity.

\subsection{Results}

\noindent\textbf{Quantitative Results.}
We compared the proposed \algname against previous state-of-the-art methods on ID preserving face generation, including FastComposer~\cite{Xiao_fastcomposer_ijcv}, PuLID~\cite{guo2024pulid}, PhotoMaker~\cite{li2024photomaker}, and a conditional SDXL reproduced based on DiffSFSR~\cite{liu2024towards} due to the absence of an open-source implementation. For models that do not support expression image inputs, we use expression text for generating facial expressions. For models without the capability to generate stylized avatars, we incorporated artistic LoRA modules.\footnote{https://civitai.com/models/124347?modelVersionId=152309}
For evaluation, each model generates 135 facial expressions for each of 100 held-out identities from the CelebA-HQ dataset.

Quantitative results are reported in Table~\ref{table:result_main}. \algname outperforms all baselines in expression accuracy (Exp. and CLIP), identity preservation (DINO and ID.), and identity consistency (DINO Con. and ID Con.).

\noindent\textbf{Qualitative Results.}
Figure~\ref{fig:qualitative_comp} presents a qualitative comparison between \algname and the two strongest baselines. 
The first row displays the reference expression image for each expression class. 
Compared to baselines, our \algname demonstrates improved identity consistency across different expressions. 
While baselines often produce variations in facial features, hairstyle, face shape, and clothing, \algname preserves the visual characteristics of the subject.
Second, \algname produces facial expressions that are more accurately aligned with the reference images, while expressions generated by baselines are inaccurate or exaggerated. 
Lastly, our method generates avatars that are more faithful to the input identity image, preserving distinctive characteristics while introducing artistic stylization. Some images generated by the baselines do not capture the key facial features of the reference identity due to cartoon stylization. Overall, \algname generates avatar images that are more expressive, identity-consistent, and faithful to the input subject compared to baseline methods. 

\noindent\textbf{Analysis on Identity Preservation.}
To further assess how well the generated avatars preserve the input identity, we conducted an additional analysis illustrated in Figure~\ref{fig:id_box_plot}.
For each of the 100 test identities, we first retrieved multiple real images of the same person from the CelebA Dataset~\cite{liu2015faceattributes}, which includes identity annotations of CelebA-HQ and multiple photos per subject.
We then computed the pairwise Euclidean distances between their normalized facial identity embeddings~\cite{deng2019arcface}, and visualized the distribution as a box plot.
This captures the natural variation in appearance for each identity across different real-world images.
Next, we calculated the average identity distance between the generated avatars and the corresponding reference identity image. The mean of these distances is shown as a red dot on the box plot. 
We observe that the red dot of our model falls within the interquartile range (Q1-Q3) and is close to the mean, indicating that the generated avatars are as consistent with the reference identity as real images of the same person. Our score also outperforms all other baselines, indicating superior identity preservation.

Additionally, we present qualitative examples of reference identities and their corresponding generated avatars in Figure~\ref{fig:id}. Expressions of the generated avatars are randomly selected. The generated avatars effectively capture the key visual characteristics of the input identity, and maintain a recognizable resemblance to the original identity.

\noindent\textbf{Alternative Styles.}
We present alternative styles including LEGO, claymation, and crayon drawing in Figure~\ref{fig:style}. These styles are achieved by incorporating curated, style-specific prompts, , \textit{e.g.} ``A LEGO-style image of a face showing a facial expression of [expression]. A playful, modular style using brightly colored, interlocking plastic bricks and simple, blocky characters with minimal detail." This highlights the flexibility and controllability of the proposed model in adapting to various artistic domains.

Lastly, an ablation study is presented in the supplementary material.
\vspace{-0.5em}
\section{Conclusions}
\vspace{-0.5em}

In this work, we present \algname, a novel framework for personalized avatar generation.
\algname generates expressive and identity-consistent avatars with fine-grained facial expressions and diverse styles.
It introduces an identity-expression conditioned multimodal diffusion transformer, building upon Stable Diffusion 3.5, which integrates identity and expression information via decoupled cross-attention, using embeddings extracted from a pretrained face encoder and an expression encoder. To further ensure consistency across generated outputs at inference, we incorporate a consistent attention mechanism that enables in-batch token sharing during generation. Quantitative and qualitative experiments demonstrate that \algname outperforms existing approaches in expression accuracy, identity fidelity, and cross-expression consistency. We also demonstrate the applicability of the framework for diverse artistic styles, highlighting its flexibility.

{
    \small
    \bibliographystyle{ieeenat_fullname}
    \bibliography{main}
}

\clearpage
\appendix

\section{Ethical Impact Statement}

Our work focuses on avatar generation with fine-grained facial expressions, aiming to support applications in education, gaming, and virtual communication. While such technologies have the potential to enrich user engagement and experience, they also raise important ethical considerations. As with any generative model, our framework carries risks of misuse. These include the potential to generate misleading or harmful content, impersonate individuals, replicate creative work without proper credit, and compromise personal privacy. Additionally, the ability to manipulate expressions may enable deceptive emotional cues or impersonation in virtual contexts.

We acknowledge these concerns and are committed to promoting the responsible development and use of generative technologies. This includes ensuring transparency, promoting ethical deployment, and explicitly discouraging the use of our model for deceptive, malicious, or privacy-infringing purposes.

\section{Ablation Study}

\begin{figure*}[t]
    \centering
    \includegraphics[width=\textwidth]{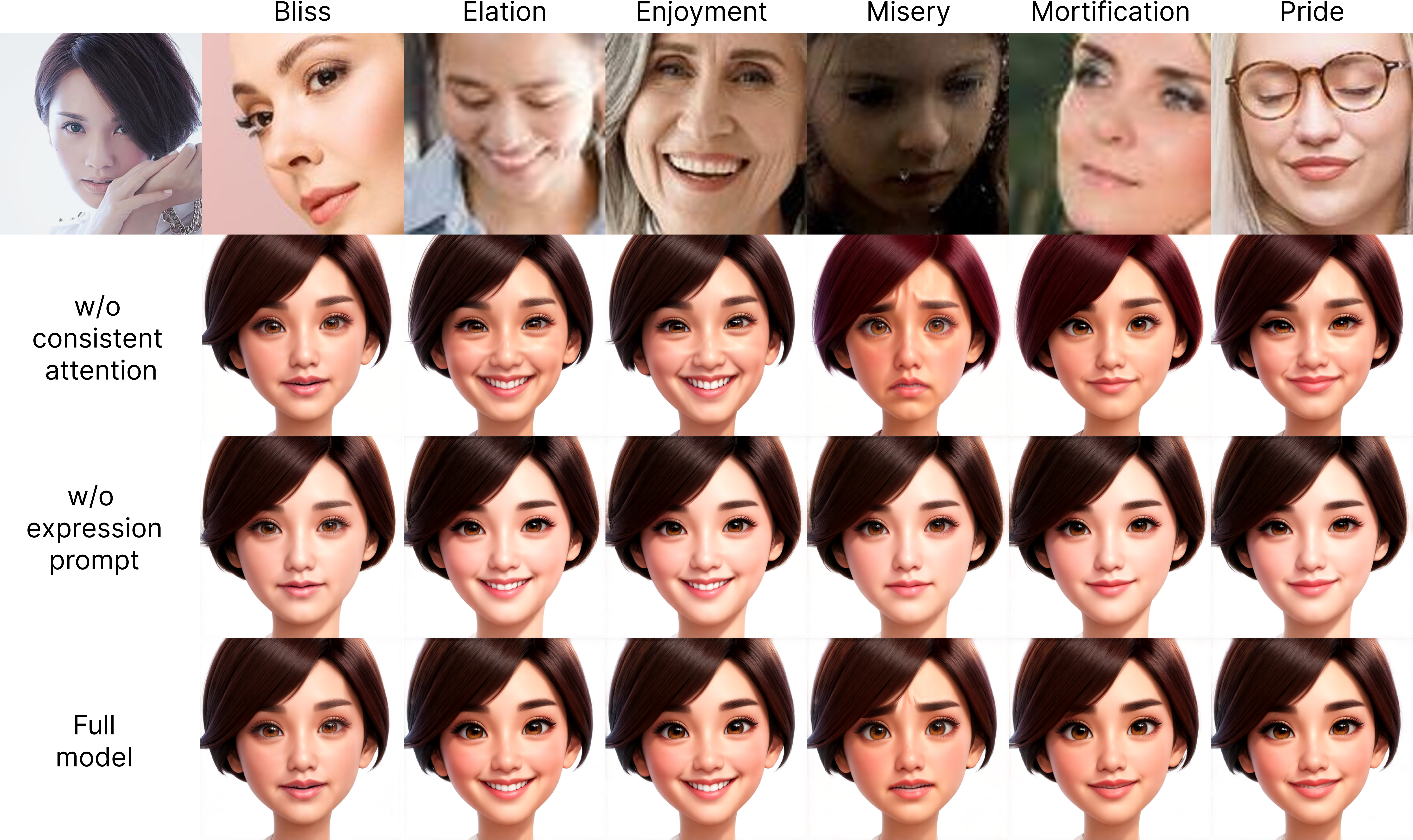}
    \caption{Qualitative results for ablation study. After removing the consistent attention module, the generated avatars exhibit inconsistencies in appearance, such as varying hair colors and noticeable differences in facial features. Removing expression text from the prompt results in less expressive generations.}
    \label{fig:ablation}
\end{figure*}

We conducted ablation studies evaluating the effects of removing the consistent attention module and removing the expression text in the input prompt (Table~\ref{tab:ablation} and Figure~\ref{fig:ablation}). We observe that removing the consistent attention module leads to a significant drop in image consistency. From the qualitative results in Figure~\ref{fig:ablation}, the generated avatars exhibit inconsistencies in appearance, such as varying hair colors and noticeable differences in facial features. Removing expression text from the prompt results in less expressive generations and lower expression-related metrics. Note that the improvement in consistency scores is due to the outputs becoming more neutral and less expressive, making them visually more similar rather than better controlled. 
Lastly, while we ablated the components above to analyze their individual impact, we chose not to ablate the three proposed loss terms individually as they are jointly essential for generating expression- and identity-controlled avatar images. The flow loss serves as the primary training objective for image generation, while the identity and expression loss ensure that the generated image reflects the intended identity and expression. Removing any of these losses results in a failure in generation. Without flow loss, the model cannot generate reasonable images. Without identity or expression loss, the model degrades into a generic T2I model with no control over identity or expression. Given this, ablating these losses would not yield meaningful insights.

\begin{table*}[t]
\centering
\begin{tabular}{lcccccc}
\hline
 & \multicolumn{2}{c}{Expression} & \multicolumn{2}{c}{Identity} & \multicolumn{2}{c}{Consistency} \\
 Model &  Exp.$\downarrow$ & CLIP$\uparrow$ & ID.$\uparrow$ & DINO$\uparrow$ & DINO Con.$\uparrow$ & ID Con.$\uparrow$ \\
\hline
Full model               & \textbf{11.09} & \textbf{0.678} & 0.361 & \textbf{0.828} & 0.957 & 0.762 \\
w/o consistent attention & 11.47 & 0.683 & 0.361 & 0.816 & 0.934 & 0.723 \\
w/o expression prompt    & 12.25 & 0.643 & \textbf{0.363} & 0.815 & \textbf{0.981} &\textbf{ 0.889} \\
\hline
\end{tabular}
\caption{Ablation study showing the effect of removing consistent attention and expression prompt.}
\label{tab:ablation}
\end{table*}

\section{Out-of-Distribution Expression Input}
\begin{figure}[t!]
    \centering
    \includegraphics[width=\columnwidth]{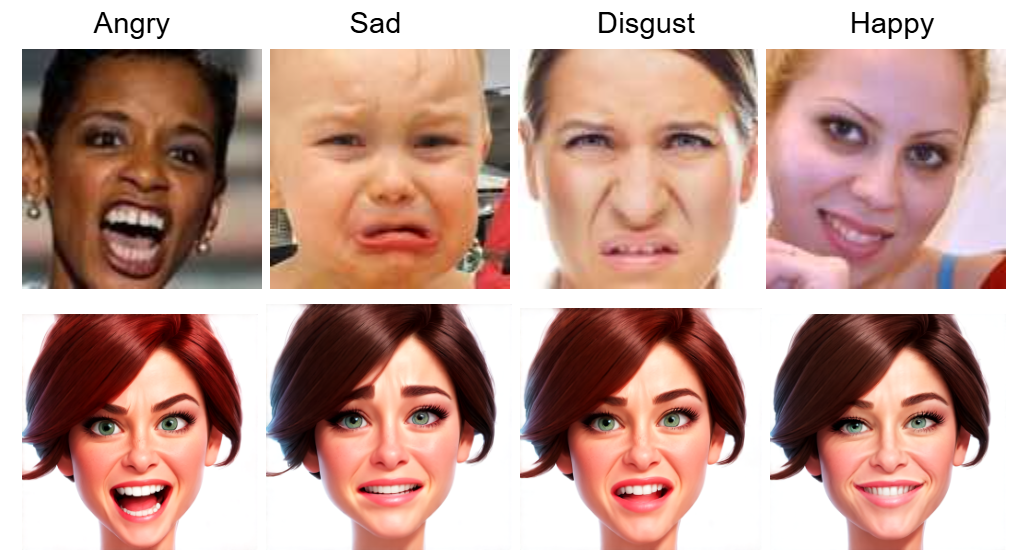}
    \caption{Generated avatar using expression images from AffectNet.}
    \label{fig:affectnet}
\end{figure}

In this work, we use expression images from Emo135 dataset~\cite{chen2022semantic} as our reference expression images. While expression images from the same dataset may yield the best performance, our model supports arbitrary expression images at inference time. To demonstrate this generalization, we tested our model with images from AffectNet~\cite{mollahosseini2017affectnet}, a dataset of basic facial expressions, and found that it can still successfully generate the corresponding expressions (Figure~\ref{fig:affectnet}). 

\end{document}